\documentclass{article}
\usepackage{spconf,amsmath,epsfig,multirow}
\usepackage{tipa}
\usepackage{cite}
\begin{document}\sloppy

\def\x{{\mathbf x}}
\def\L{{\cal L}}

\title{DEEP SEGMENTAL PHONETIC POSTERIOR-GRAMS BASED DISCOVERY OF NON-CATEGORIES IN L2 ENGLISH SPEECH}
%
\name{Xu Li, Xixin Wu, Xunying Liu, Helen Meng}


\address{Department of Systems Engineering and Engineering Management, \\
                   The Chinese University of Hong Kong \\
{\small \tt \{xuli, wuxx, xyliu, hmmeng\}@se.cuhk.edu.hk}}

\maketitle

\begin{abstract}
Second language (L2) speech is often labeled with the native, phone categories. However, in many cases, it is difficult to decide on a categorical phone that an L2 segment belongs to. These segments are regarded as non-categories. Most existing approaches for Mispronunciation Detection and Diagnosis (MDD) are only concerned with categorical errors, i.e. a phone category is inserted, deleted or substituted by another. However, non-categorical errors are not considered. To model these non-categorical errors, this work aims at exploring non-categorical patterns to extend the categorical phone set. We apply a phonetic segment classifier to generate segmental phonetic posterior-grams (SPPGs) to represent phone segment-level information. And then we explore the non-categories by looking for the SPPGs with more than one peak. Compared with the baseline system, this approach explores more non-categorical patterns, and also perceptual experimental results show that the explored non-categories are more accurate with increased confusion degree by 7.3\% and 7.5\% under two different measures. Finally, we preliminarily analyze the reason behind those non-categories.
\end{abstract}
\section{Introduction}
\label{sec:intro}

Mispronunciation Detection and Diagnosis (MDD) is one of the core technology in Computer-Aided Pronunciation Training (CAPT) systems. It aims to detect the mispronunciation in a second language (L2) learner's speech, and further diagnose the error type and give learners effective feedback. There are several ways to address this task: The methods based on pronunciation scoring \cite{franco1997automatic,witt2000phone,zheng2007generalized,hu2013new} are very popular and achieve a promising performance on mispronunciation detection. However, this kind of methods can not deal with mispronunciation diagnosis. Alternative methods, such as extended recognized network (ERN) \cite{harrison2009implementation,lo2010automatic,qian2010discriminative} and acoustic-phonemic model (APM) \cite{li2017mispronunciation}, also perform well. ERN incorporates manually designed or data-derived phonological rules to generate possible phone paths in a word, including a canonical phonemic path and common mispronunciation paths. If a decoded path involves a mispronunciation path, then the mispronunciation can be detected and also diagnosed. Different from traditional acoustic models, APM feeds in not only acoustic features, but also phone context information for better performance.

\begin{figure}[t]
  \centering
  \includegraphics[width=\linewidth]{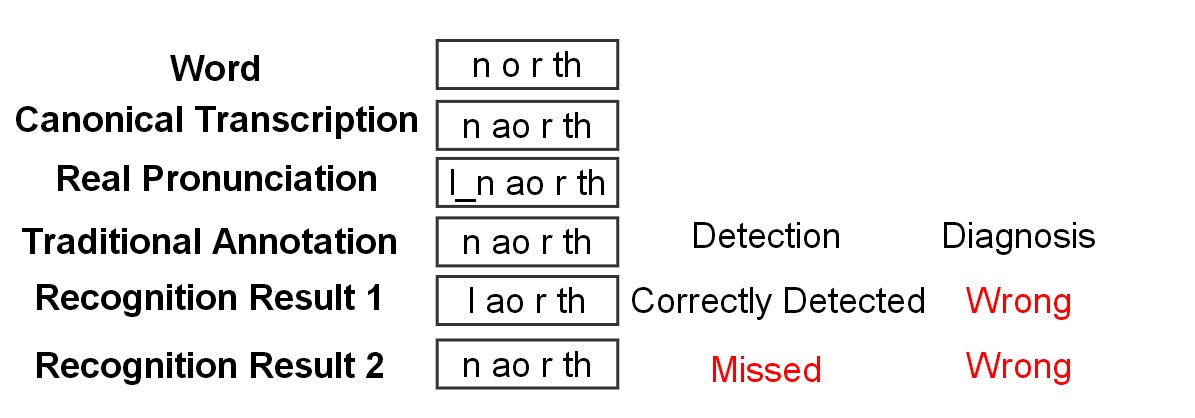}
  \caption{An example for how non-categorical mispronunciations are wrongly treated in traditional MDD}
  \label{fig: Example for non-categories}
\end{figure}

Based on the categorical phone set, most existing approaches to model L2 speech only target categorical phone errors, but ignore non-categorical errors (i.e., the segments for which it is difficult to label as a single phone category). For example, L2 English speech uttered by native Cantonese speakers shows that the phone [n] is often pronounced as a sound that bears resemblance to both [l] and [n] ([l\_n] in Fig.~\ref{fig: Example for non-categories}). In current MDD proposals, they are often coarsely labeled as one of the approximate phones. Fig.~\ref{fig: Example for non-categories} shows an example where the canonical annotation for "north" should be [n ao r th], but in face of the non-categorical pronunciation that resembles both [l] and [n], it may be recognized as either [l ao r th] (as in Recognition Result 1) which enables mispronunciation detection but inaccurate diagnosis. Alternatively, if the non-categorical segment is recognized as [n ao r th] (as in Recognition Result 2), it will fail to neither detect nor diagnose the mispronunciation. From the learner's perspective, the unawareness of non-categorical errors resulting from the training system will greatly decrease the efficiency of their learning, and they may even pronounce with their own accents through the training.

\begin{figure*}[t]
  \centering
  \includegraphics[width=\linewidth]{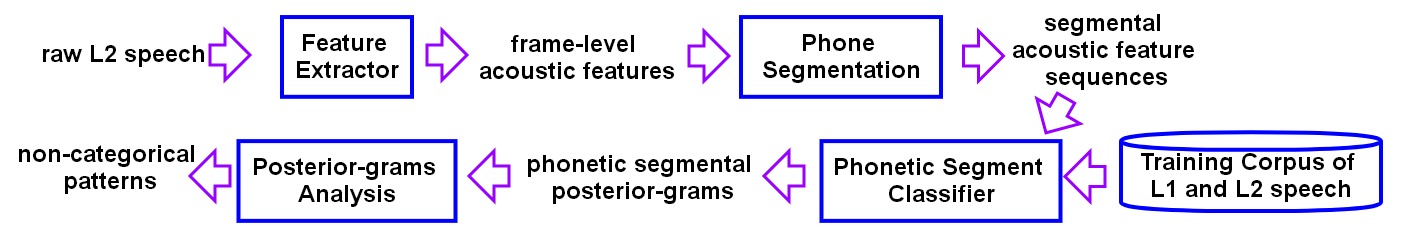}
  \caption{Non-categorical patterns acquisition framework}
  \label{fig: framework}
\end{figure*}

Concerned with the above problem, there are several approaches \cite{wang2015supervised,lee2016personalized,mao2018unsupervised,wang2013toward,li2018unsupervised} proposed to capture the L2 pronunciation deviations from categorical phones. \cite{wang2015supervised,li2018unsupervised,wang2013toward} try to capture each L2 error patterns of a given categorical phone, while \cite{mao2018unsupervised} focuses on extending a native phone set to include the non-categorical patterns in L2 English speech. Actually, this work is an extension of \cite{mao2018unsupervised}. The distinction lies in 1) this approach extracts segment-level features to model the phonetic information while \cite{mao2018unsupervised} uses non-segmental frame-level features and their average, thereby the accumulated errors are reduced and the discovered non-categories are more accurate with higher confusion degree; 2) this work uses a more simple but effective method to explore non-categories, while not involving k-means clustering method used in \cite{mao2018unsupervised}. This prevents a data imbalance problem and high-dimension data issues in k-means clustering, and hence discovers more non-categories.

The contributions of this work include 1) introduce a novel non-categorical patterns acquisition framework; 2) preliminarily analyze the reasons behind these non-categories.

The rest of this paper is as follows. In Section 2, we describe the framework and approaches. Experiment setups and results are presented in Section 3 and some analysis for these discovered non-categories are conducted in Section 4. Finally, Section 5 concludes the paper.

\section{Framework and approaches}

The framework in this work is shown in Fig.~\ref{fig: framework}. We first extract acoustic features of the raw speech on frame level. And then, since the text transcription is available, forced alignment is performed to concatenate acoustic features into segment-level features corresponding to phonemes. Feed on these segmental acoustic feature sequences, a phonetic segment classifier is trained on the corpus of both L1 and L2 speech. We use output of the softmax layer, i.e. segmental phonetic posterior-grams (SPPGs) to represent phone segment-level information. This representation is widely used in the L2 English acoustic-phonemic space \cite{wang2015supervised,sun2016phonetic,wang2013toward,lee2013mispronunciation} due to its speaker independence, and more environment-robust properties. Based on SPPGs, an effective method is used to explore non-categories.

\subsection{Phonetic segment classifier}

\begin{figure*}[t]
  \centering
  \includegraphics[width=0.8\linewidth]{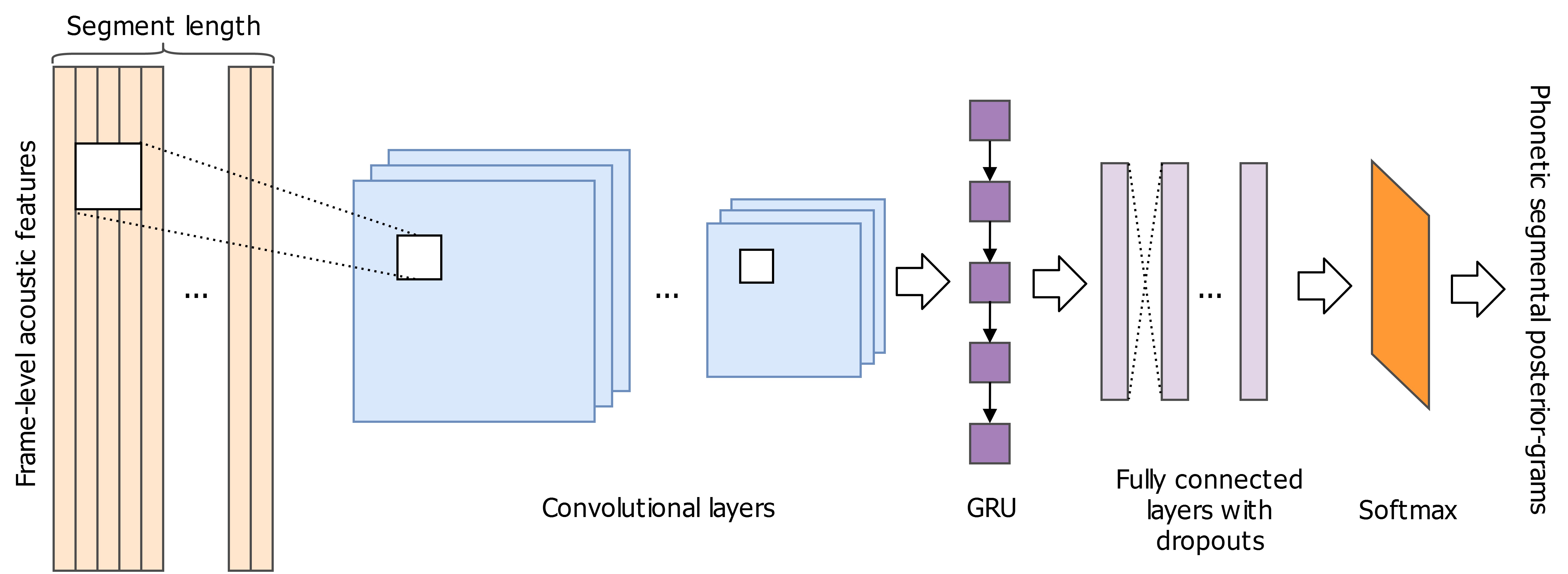}
  \caption{Phonetic segment classifier architecture}
  \label{fig: Model Architecture}
\end{figure*}

This phonetic segment classifier consists of convolutional layers, recurrent layers and dense layers, as shown in Fig.~\ref{fig: Model Architecture}. This unified architecture \cite{sainath2015convolutional} takes advantages of the complementarity of Convolutional Neural Networks (CNNs), Recurrent Neural Networks (RNNs) and Deep Neural Networks (DNNs), and it is widely used in Automatic Speech Recognition (ASR) tasks \cite{amodei2016deep,xiong2018microsoft,trigeorgis2016adieu}. In this work, we applied 2D CNN upon the input segmental acoustic feature sequences. The output of the final CNN layer is fed to the RNN layer. Since Gated Recurrent Unit (GRU) \cite{cho2014learning} has a performance on par with Long Short-Term Memory (LSTM) \cite{hochreiter1997long} while is more efficient in computation, we adopt GRU layer in our model. The final state of the GRU layer is fed to the FC layers and the softmax layer. Within this structure, CNNs serve as the speaker-independent feature extractor, which is robust to spectral variation. RNNs can model the temporal characteristics within a phone segment. DNNs are used to transform the final hidden features into a space that makes the output easier to classify.


To train this phonetic segment classifier, we need L1 (native) speech to implicitly guide the model to learn what canonical pronunciation of a phone should be. At the same time, since second language (L2) space has some deviations from the native space, we also need L2 speech to adapt the model parameters to fit it well. Hence both L1 and L2 speech corpus are used for training this classifier. In Section 3, the segment-level phone recognition rates are shown under different training corpus setups, and it indicates the importance and correctness of using both L1 and L2 speech to train the model.

\subsection{Non-categories exploration}
The output of the classifier is segmental phonetic posterior-grams, which is a probability vector where each element corresponds to a phone category. The higher probability one element preserves, the more likely the segment belongs to the phone category which the element corresponds to. According to this, for a segment of canonical pronunciation, the classifier will output a vector with a high probability (confidence) on that canonical phone element. However, for the segments of L2 non-categories, the classifier will be ambiguous on more than one element, which results in a probability vector with more than one peak of comparable probabilities. From this point, we can set a threshold and look for the segments where the probability vector has more than one element higher than this threshold. These segments can be regarded as non-categories. After this exploration, we can use an identical name to denote all the segments with the same element set above the threshold. Then we have non-categorical patterns. For example, we can use $l\_n$ to represent the non-categorical segments with an element set of \{l, n\} above a given threshold. Section 3 gives the perceptual test results on these explored non-categories.

\begin{table}
\begin{center}
\caption{Experimental data setup} \label{tab:data setup}
\begin{tabular}{c|c|c}
  \hline
                & training set & evaluation set
  \\
  \hline
  TIMIT    & 3.0 hours & 0.4 hours \\
 (L1)      & (144K segments) & (18K segments) \\
  \hline
CU-CHLOE-C & 5.5 hours & 0.7 hours \\
   (L2)   & (139K segments) & (19K segments) \\
  \hline
\end{tabular}
\end{center}
\end{table}

\section{EXPERIMENTS}

\subsection{Experimental setup}

Our experiments are conducted on two speech corpora: 1) the CU-CHLOE-C (Chinese University-Chinese Learners of English - Cantonese speakers) data set \cite{meng2010development} as the L2 speech corpus; and 2) the TIMIT data set as the L1 speech corpus. Both L1 and L2 corpus are divided by speakers into training set and evaluation set, as shown in Table~\ref{tab:data setup}.

The acoustic feature adopted in this work is 13 dimensional Mel-Frequency Cepstrum Coefficient (MFCC). There are 3 CNN layers with 64 output channels for each. The kernel size is 3 by 3, and no pooling function is used. The GRU layer has 128 hidden units. Its final state is transformed to output by 3 dense layers with 512 hidden units per layer, and the dropout \cite{srivastava2014dropout} rate is set as 0.2 to prevent over-fitting problems. The implementation software used in this work is Pytorch 0.4.1 \cite{Ketkar2017}. We use the combined training set of L1 and L2 to train the model, and use the evaluation set to test the performance on each corpus. During the training, we split 10\% data from training set as the validation set. Adam \cite{kingma2014adam} is used as the optimizer, and the initial learning rate is set as 0.001. The probability threshold used for exploring non-categories is set to 0.4.


\begin{table}[t]
\begin{center}
\caption{Segment-level phone recognition rate} \label{tab:recognition results}
\begin{tabular}{c|c|c|c}
  \hline
  training & training & evaluation set & evaluation set \\
  set & set & (L1) & (L2) \\
  \hline
  L1 & 74.27\% & 72.92\% & 32.12\% \\
  \hline
  L2 & 80.08\% & 27.24\% & 71.02\% \\
  \hline
  L1+L2 & 75.24\% & 72.06\% & 70.29\% \\
  \hline
\end{tabular}
\end{center}
\end{table}

\begin{table*}[t]
\begin{center}
\caption{Average proportion of four options in each non-categories (The previous results \cite{mao2018unsupervised} are shown in parentheses). The higher score of Option 3 and lower score gap between Option 1 and 2 are preferable.} \label{tab: average results}
\begin{tabular}{ccccccccccc}
  \hline
                            & [ax\_er]& [aa\_ao]& [ey\_ih]& [ae\_ay]& [eh\_ey]& [d\_t]  & [l\_n]  & [b\_p]  & [f\_v]  & [m\_n] \\
  \hline
  \multirow{2}{*}{Option 1} & 38.9\% & 35.2\% & 56.5\% & 25.9\% & 14.7\% & 49.0\% & 24.1\% & 27.8\% & 51.9\% & 41.1\% \\
                            &(46.0\%)&(16.1\%)&(46.6\%)&(25.0\%)&(17.7\%)&(57.8\%)&(56.6\%)&(18.8\%)&(37.2\%)&(55.6\%) \\
  \hline
  \multirow{2}{*}{Option 2} & 37.0\% & 48.1\% & 25.0\% & 44.4\% & 63.7\% & 28.7\% & 43.5\% & 35.2\% & 29.6\% & 36.3\% \\
                            &(30.6\%)&(50.0\%)&(27.8\%)&(47.5\%)&(60.4\%)&(18.0\%)&(21.7\%)&(35.2\%)&(45.7\%)&(27.5\%) \\
  \hline
  \multirow{2}{*}{Option 3} & \textbf{13.0\%} & 9.3\%  & \textbf{13.9\%} & \textbf{14.9\%} & \textbf{19.6\%} & \textbf{16.7\%} & \textbf{12.0\%} & \textbf{15.7\%} & \textbf{14.8\%} & \textbf{15.7\%} \\
                            &(3.2\%) &(\textbf{9.7\%}) &(4.4\%) &(7.3\%) &(3.1\%) &(7.0\%) &(8.0\%) &(8.6\%) &(11.0\%)&(8.8\%) \\
  \hline
  \multirow{2}{*}{Option 4} & 11.1\% & 7.4\%  & 4.6\%  & 14.8\% & 2.0\%  & 5.6\%  & 20.4\% & 21.3\% & 3.7\%  & 6.9\% \\
                            &(20.2\%)&(24.2\%)&(21.2\%)&(20.2\%)&(18.8\%)&(17.2\%)&(13.7\%)&(37.4\%)&(6.2\%) &(8.1\%) \\
  \hline
  \multirow{2}{*}{Score gap} \footnotemark[1] & \textbf{1.9\%} & \textbf{12.9\%} & 31.5\% & \textbf{18.5\%} & 49.0\% & \textbf{20.3\%} & \textbf{19.4\%} & \textbf{7.4\%} & 22.3\% & \textbf{4.8\%} \\
   & (15.4\%) & (33.9\%) & (\textbf{18.8\%}) & (22.5\%) & (\textbf{42.7\%}) & (39.8\%) & (34.9\%) & (16.4\%) & (\textbf{8.5\%}) & (8.1\%) \\
   \hline
  \multicolumn{11}{r}{Option 1: More similar to $P_1$; Option 2: More similar to $P_2$;} \\
  \multicolumn{11}{r}{Option 3: Equal similarity to $P_1$ and $P_2$; Option 4: Not similar to either $P_1$ or $P_2$.} \\
\end{tabular}
\end{center}
\end{table*}

\subsection{Experiment results}
Table~\ref{tab:recognition results} shows the phone recognition performance of our system trained on different training sets. If only L1 corpus involved in training, the system can have a phone recognition rate of 72.92\% and 32.12\% on L1 and L2 evaluation set respectively. This implies there is some deviation from L2 space to L1 space. On the counterpart, if only L2 corpus involved in training, the system has a phone recognition rate of 27.24\% and 71.02\% on L1 and L2 evaluation set respectively. However, if a combined training set of both L1 and L2 is used, the performance on L1 and L2 evaluation set can almost respectively reach the phone recognition rate when merely using the corresponding corpus as the training set. This implies even though there is some deviation between L1 and L2 space, a model trained on a combination training set can project them into a common space and still preserve the recognition performance. This system will be used for non-categories exploration.

\subsection{Perceptual tests}
To verify the explored non-categories indeed perceptually different from their corresponding categorical phones, we designed perceptual tests on series of non-categorical confusion groups. Each non-categorical confusion group consists of audios from the non-category, and also the corresponding categorical phones. In our case, since the threshold to explore non-categories is set to 0.4, we only consider the non-categories that resemble between two categories. For simplified writing, we mark the non-category that resemble two categories $P_1$ and $P_2$ as $P_1$\_$P_2$.

In each group, we randomly select 12 audios (6 audios from the non-category, 3 audios from each of two categories). For the audios in each category, we select the segments where the SPPGs have a high probability (confidence) of 0.9 at that phone element.

These perceptual tests are conducted in the same way as \cite{mao2018unsupervised}. During the tests, the listeners are asked to give their perceptual choice among 4 options after carefully listening each segment with short contexts: 1) More similar to $P_1$; 2) More similar to $P_2$; 3) Equal similarity to $P_1$ and $P_2$; 4) Not similar to either $P_1$ or $P_2$. 25 listeners majoring in linguistics or English were invited to participate in these perceptual tests.

\begin{figure}
  \centering
  \includegraphics[width=\linewidth]{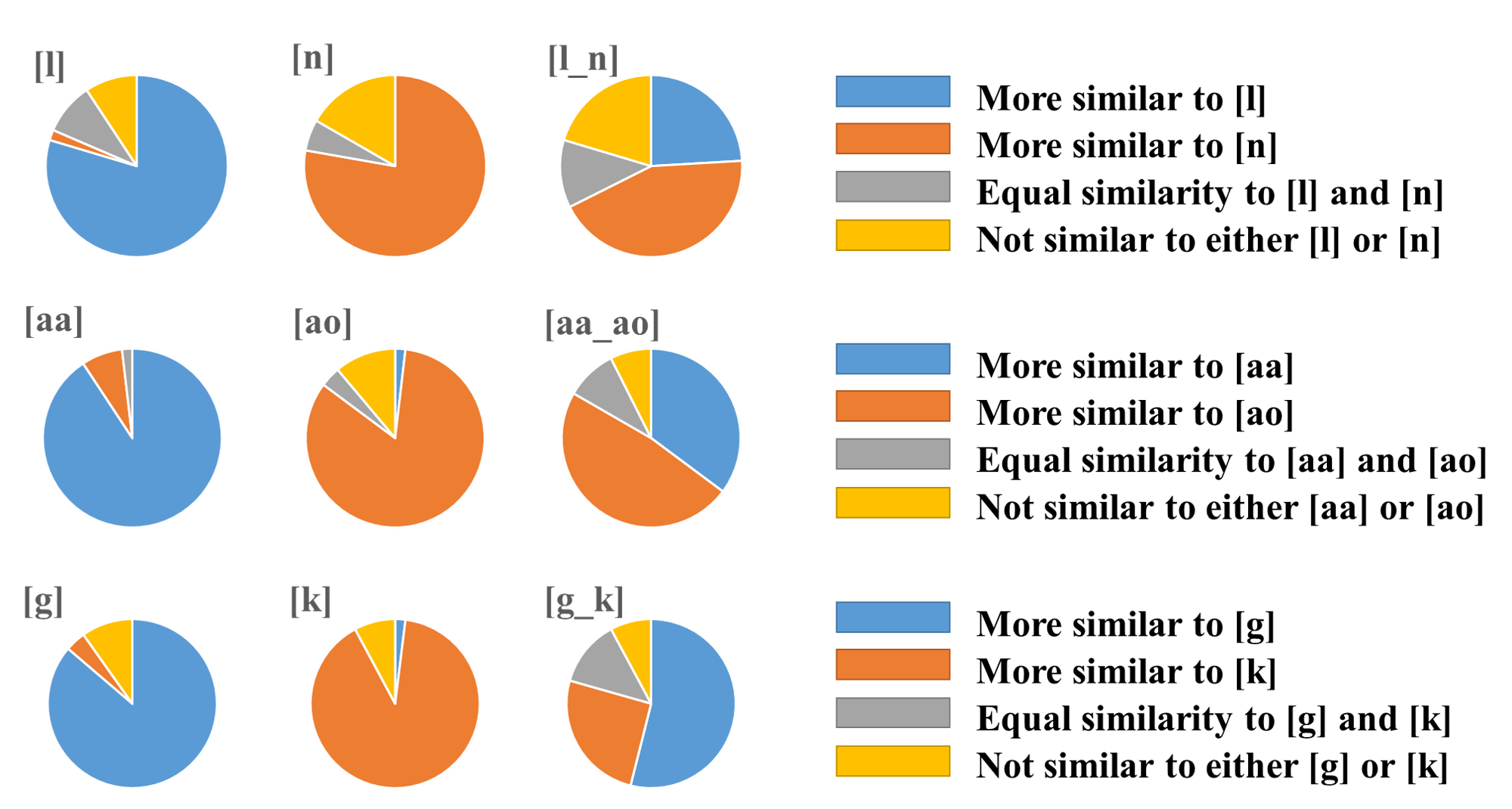}
  \caption{Statistic perceptual results of each non-categorical confusion group}
  \label{fig: statistic results of each non-category group}
\end{figure}

Fig.~\ref{fig: statistic results of each non-category group} shows the average proportion of 4 options in three sample non-categorical confusion groups. The complete confusion scores are provided in Supplemental Material A. From these pie charts, we observe that listeners almost have an agreement on native phone categories, and give the corresponding phones as their choices. However, for non-categories, there is no majority perceptual vote among the given categorical phones. This implies these non-categories are indeed perceptually different from the corresponding categories, and bear a resemblance to both of them. \footnotetext[1]{The score gap between Option 1 and 2}

To compare the confusion degree of the explored non-categories in \cite{mao2018unsupervised} and this work, we display the averaged confusion scores of the non-categories that are discovered in both \cite{mao2018unsupervised} and this work. Table~\ref{tab: average results} shows the average proportion of each option in such non-categories. The averaged confusion scores in this work are shown in the first line of each option, while the previous results are shown within parentheses in the second line. The score gap between Option 1 and 2 is worked out and displayed in the bottom part. We observe that for most non-categories, such as [ax\_er], [aa\_ao] and [d\_t], the score gap of the first two options is much smaller in this work. From the average sense, this score gap is decreased from 26.1\% to 18.8\%. Moreover, the averaged score of Option 3 is also obviously increased from 7.1\% to 14.6\%. These two observations verify that the non-categories explored in this approach are more accurate with higher confusion degree.

Table~\ref{tab:noncategories comparison} shows the explored non-category set compared with that in \cite{mao2018unsupervised}. There are 9 additional and 10 existing non-categories. To explore the possible reason, the k-means clustering algorithm used in \cite{mao2018unsupervised} may be subjected to a data imbalance problem, since the non-categorical data, especially for some non-categories (for e.g., r\_w), is far less than the categorical data. This imbalance data problem makes it difficult to cluster out those non-categories of a small amount. In contrast to this, this approach does not involve the whole data structure for clustering, so not suffer from this imbalance data problem. However, there are 2 non-categories absent from this explored set. Actually, they are observed, but the amount of these patterns is too small to be investigated.

The sample audios for native phone categories and those explored non-categories can be found in this link: https://anonymousdemos.github.io/ICME2019demos.github.io/.

\begin{table}
\begin{center}
\caption{The explored non-category set comparison between this work and the previous work \cite{mao2018unsupervised}.} \label{tab:noncategories comparison}
\begin{tabular}{ccc|ccc|c}
  \hline
  \multicolumn{3}{c|}{explored additional} & \multicolumn{3}{|c|}{explored but existing} & \multirow{2}{*}{missing} \\
  \multicolumn{3}{c|}{non-categories} & \multicolumn{3}{|c|}{non-categories} & \\
  \hline
  ah\_ax & ax\_ix & ih\_ix & ax\_er & aa\_ao & ey\_ih & aa\_ax \\
  er\_r  & ch\_t  & g\_k   & eh\_ey & ae\_ay & d\_t   & aw\_ax \\
  r\_w   & dh\_l  & s\_z   & l\_n   & b\_p   & f\_v   & \\
         &        &        & m\_n   &        &        & \\
  \hline
\end{tabular}
\end{center}
\end{table}

\begin{table}
\begin{center}
\caption{Non-categories caused by articulation absence from L1 language. The absence types are adapted from \cite{meng2007deriving}.} \label{tab:non-categories caused by articulation}
\begin{tabular}{c|ccc}
  \hline
   & \multicolumn{3}{|c}{non-categories} \\
  \hline
  Missing voiced plosives & b\_p  & d\_t & g\_k \\
  \hline
  Missing affricates & \multicolumn{3}{c}{ch\_t} \\
  \hline
  Missing fricatives & f\_v & dh\_l & s\_z \\
  \hline
  Missing and confused approximants & \multicolumn{3}{c}{r\_w} \\
  \hline
  Confusion among [l] and [n] & \multicolumn{3}{c}{l\_n} \\
  \hline
\end{tabular}
\end{center}
\end{table}

\section{Some analysis for the explored non-categories}
Based on the knowledge of non-categories existence, this section tries to analyze the reason behind it. \cite{meng2007deriving} derived salient learners' mispronunciations from cross-language phonological comparisons. Inspired by this, L1 phonetic transfer is regarded as the main reason behind these non-categories.

\subsection{Vowels and diphthongs}
\begin{figure}
	\centering
	\begin{minipage}[b]{0.6\textwidth}
		\includegraphics[width=0.4\textwidth]{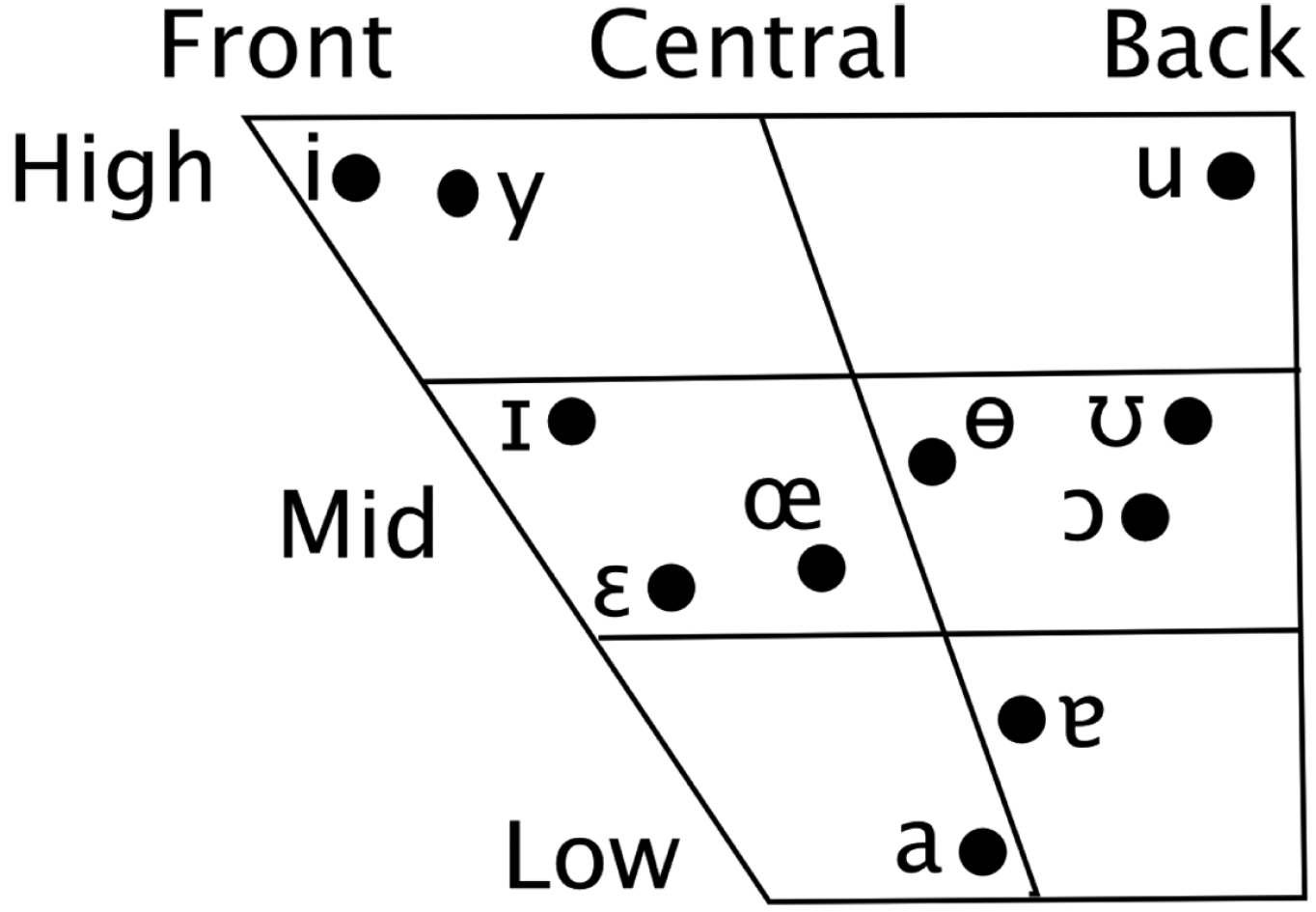}
		\includegraphics[width=0.4\textwidth]{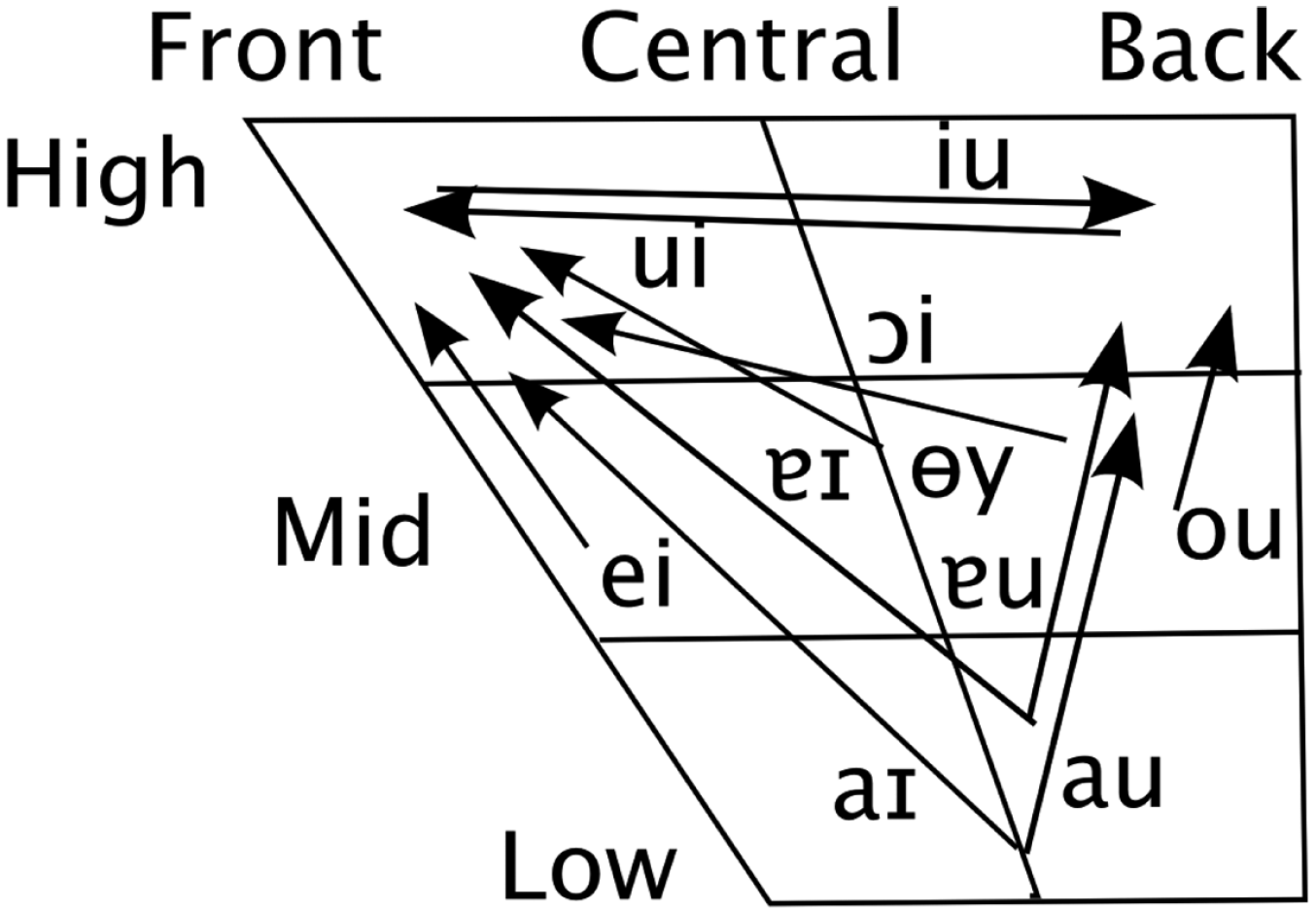}
	\end{minipage}
	\caption{Cantonese vowels and diphthongs \cite{zee1991chinese}. Tongue positions (front, central, back, high, mid, low) are labeled.}
	\label{fig: Cantonese vowels}
\end{figure}

\begin{figure}
	\centering
	\begin{minipage}[b]{0.6\textwidth}
		\includegraphics[width=0.4\textwidth]{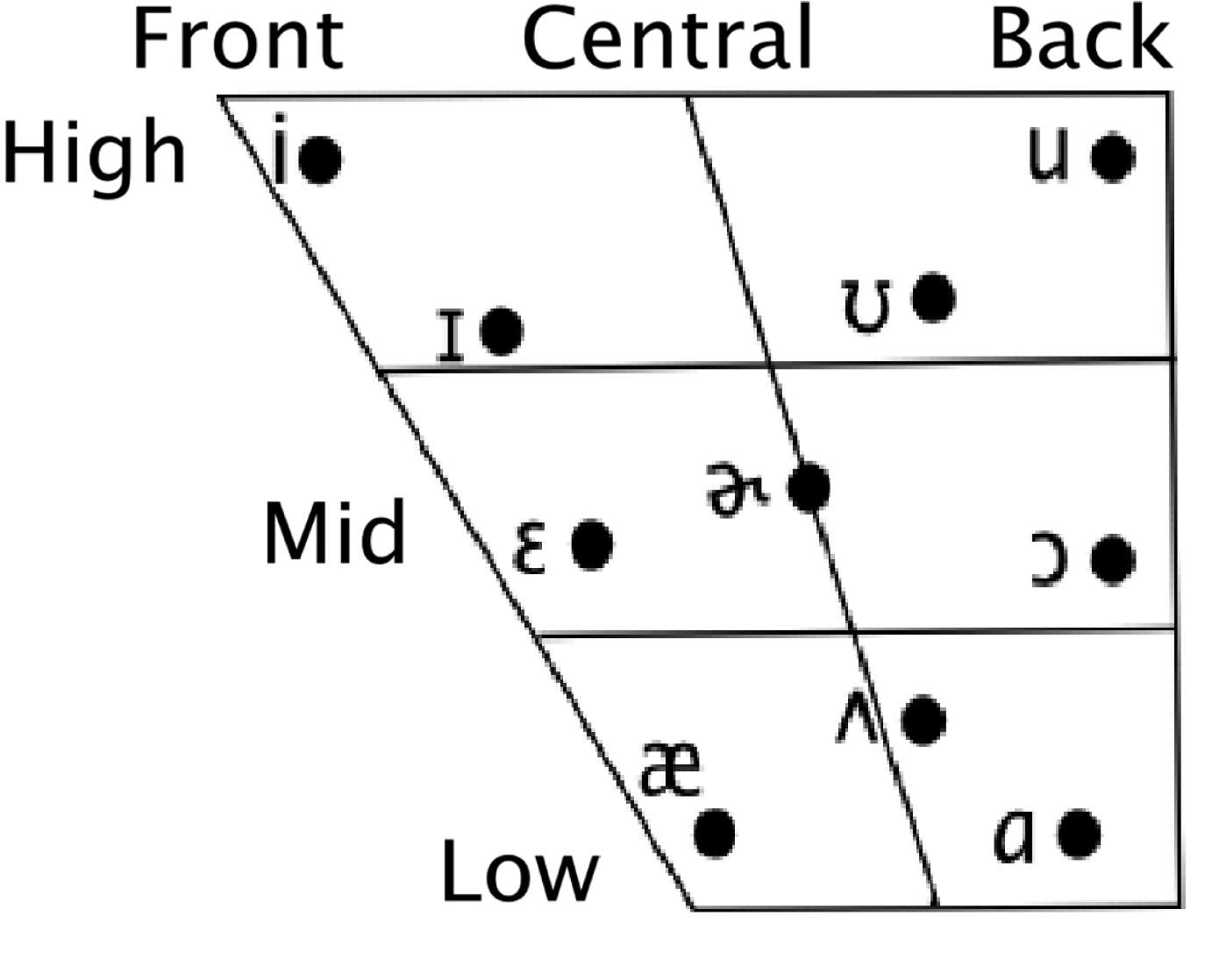}
		\includegraphics[width=0.4\textwidth]{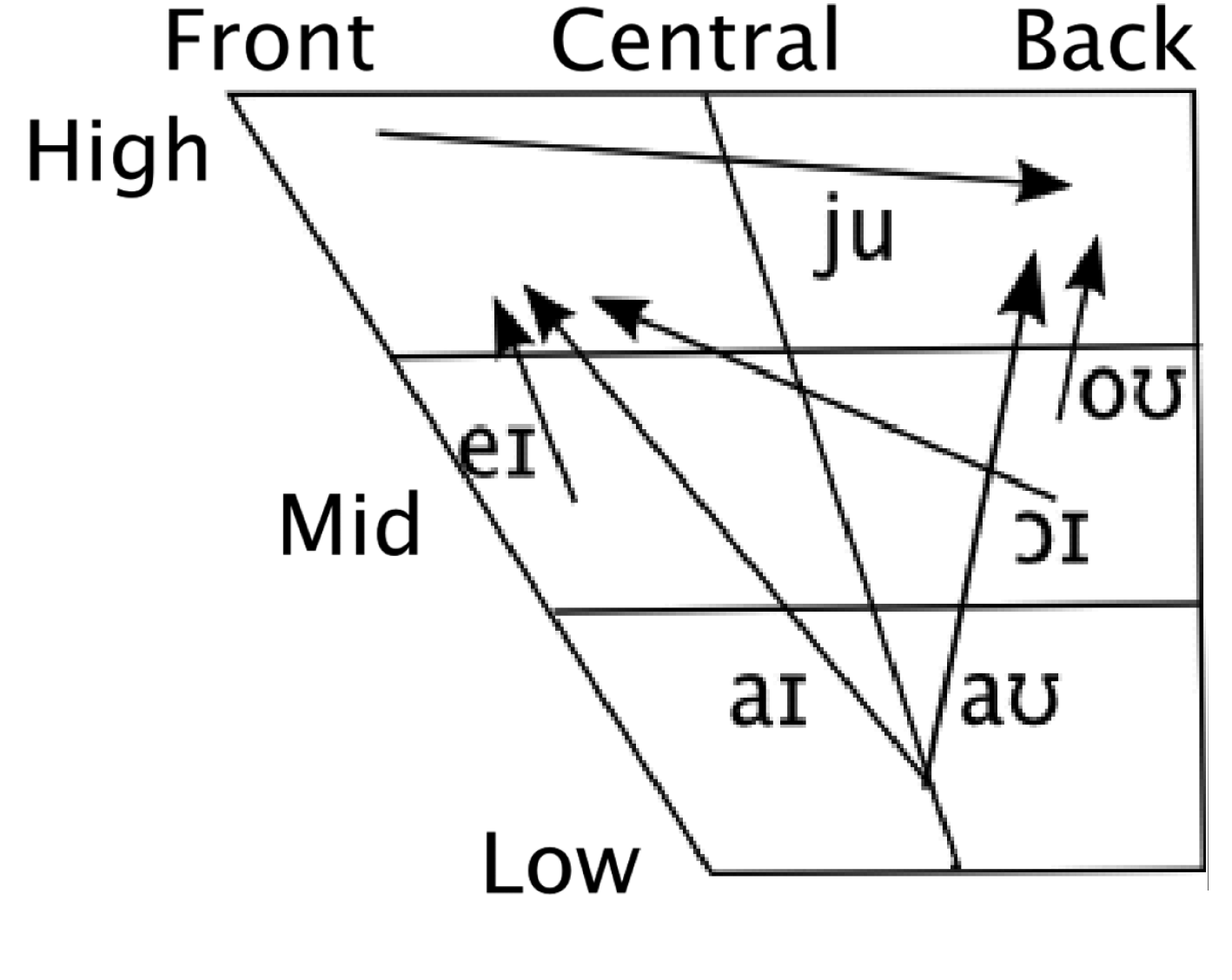}
	\end{minipage}
	\caption{American English vowels and diphthongs \cite{decker1999handbook}. }
	\label{fig: American English vowels}
\end{figure}

In this work, the learners' L1 language is Cantonese, which has a different phone inventory compared with English. The Cantonese vowel inventory is generally richer than English, however it lacks the rhotic and many low vowels found in English \cite{meng2007deriving}. The vowel inventories \footnote[2]{These vowels are denoted in International Phonetic Alphabet (IPA) form.} of Cantonese and English are illustrated in Fig.~\ref{fig: Cantonese vowels} and Fig.~\ref{fig: American English vowels} respectively. A comparison of these two inventories shows that the English vowel [ae] ([\textipa{\ae}]), [ax] ([\textipa{@}]), [ah] ([\textipa{2}]) and [aa] ([\textipa{A}]) are missing in Cantonese. The lack of these vowels in the learner's L1 is hypothesized to be a source of possible mispronunciation in L2 \cite{lado1957linguistics}, and L2 learners tend to pronounce these missing vowels from their L1 which are phonetically-similar to them (L1 transfer). This tendency makes the pronunciation of these vowels not accurate, and produces the non-categories, such as [ah\_ax], [aa\_ao] and [ae\_ay].

\subsection{Consonants}
Similar observations exist in consonant inventories comparison of Cantonese and English. Due to the limitation of space, the consonant inventories of English and Cantonese are put in the Supplemental Material B. \cite{meng2007deriving} analyzed the English consonants missing in Cantonese and divided them into classes according to different articulation absence. These absence may also result in the corresponding non-categories, as shown in Table~\ref{tab:non-categories caused by articulation}.

There are three English voiced plosives ([b], [d] and [g]) missing from Cantonese, and often being pronounced similarly with the voiceless, unaspirated Cantonese plosives ([p], [t] and [k]). This results in the non-categories of [b\_p], [d\_t] and [g\_k]. And also, [ch] is an English affricate missing in Cantonese, and it is often replaced with the aspirated alveolar Cantonese affricate [$ts^h$], which bears a resemblance to both English [ch] and [t]. Similarly, [f\_v], [dh\_l] and [s\_z] are caused by the missing English fricatives [v], [dh] and [z] in Cantonese respectively. [r\_w] is produced by the missing English approximant [r], which is often substituted with [w] or [l] in Cantonese. Finally, it is often acceptable to substitute [n] with [l] in colloquial Cantonese, for e.g. 你(you) [nei] pronounced as 理(logic) [lei] \cite{meng2007deriving}. Hence, Cantonese learners often unconciously perform these substitutions in English, resulting in the non-category [l\_n].

\section{CONCLUSIONS}
This work introduces a novel non-categorical phone acquisition framework, based on SPPGs learned via deep neural networks. A phonetic segment classifier is utilized to extract SPPGs, and we apply a simple but effective method to explore the potential non-categories. This approach not only helps increase the confusion degree of the explored non-categories, thereby improving the accuracy of non-categories exploration, but also discovers more non-categorical patterns. Possible reasons behind the non-categories are also given based on the phone inventories comparison between L1 and L2 languages. Since this framework has not language specific constraint, it can be smoothly adapted to other language pairs. In the future, how to facilitate MDD performance with these non-categories will be investigated.

\bibliographystyle{IEEEbib}
\small{\bibliography{template}}

\end{document}